\documentclass[11pt]{article}
\usepackage[T1]{fontenc}
\usepackage[utf8]{inputenc}
\usepackage[english]{babel}
\usepackage[margin=1in]{geometry}
\usepackage{lmodern,microtype}
\usepackage{amsmath,amssymb,amsthm,booktabs,graphicx,longtable,array}
\usepackage[font=small,labelfont=bf]{caption}
\usepackage[numbers]{natbib}
\usepackage{enumitem}\setlist{itemsep=1pt,topsep=3pt,parsep=0pt}
\usepackage{hyperref,cleveref,orcidlink,titling,fancyhdr}

\newcommand{\KL}{\operatorname{KL}}
\newcommand{\Lcat}{\ensuremath{\mathrm{L}}}
\newcommand{\Ncat}{\ensuremath{\mathrm{N}}}
\newcommand{\Hcat}{\ensuremath{\mathrm{H}}}

\newif\ifpaperhasdoi
\paperhasdoifalse
\newcommand{\manuscriptversion}{Preprint v1.0}
\newcommand{\manuscriptdate}{2 October 2026}
\newcommand{\manuscriptyear}{2026}

\expandafter\def\csname captionfig1\endcsname{\textbf{Two task layers and their history witness.} (a) A maps four records to five per-slot sums. B uses their slot-1 category and the preceding mode to update the mode and determine the next-category law. The annotations identify training conditions, not evidence that a network learned the depicted decomposition. (b) The histories end on the same neutral sum board but retain different modes and require different laws. Values are integer twelfths; law entries are ordered L/N/H. The task has 24,435 completed sum boards and 1,555 local histories. Everything shown is exact task ground truth, not model output.}
\expandafter\def\csname captionfig2\endcsname{Recoverability, computation and access are distinct measurements, not interchangeable routes. (a) The sums output head and its loss attach to the shared carrier; a dashed optional connection feeds sums output to prediction in the connected variants. (b) The matrix identifies the routes and losses in the three versions and r13 variants; probability targets retain their base version's routes and replace sampled targets with the exact law. The probe fits a reader on frozen states; the swap replaces an aligned donor block, with selectivity checked separately. No measurement, seed average or control result is implied by the icons; procedures and their calibration are in Appendix C.}
\expandafter\def\csname captionfig3\endcsname{Recoverable sums and low average error coexist with specific failures. (a,b) r8 linear/MLP; (c) r10 linear: slots 1–5, 512 held-out boards/slot, majority ticks, shuffled floors and oracle ceiling. Open/filled: update 0/20k (a–c), r10/r8 (d,e); shapes: seeds 0/1/2. (d) Natural KL: 60,000 positions, board-only/mode-blind floors; the log axis omits the oracle's 0 bits. (e) Law TV: 576 r8 histories/180 r10 rows, zero oracle, r8 bar 0.02 and r10 original-law 0.05 bar (its equal-law bar is 0.02). (f) r12 uniform categorical misreads: 24,435 boards, rendering 0, untrained floor, shaded cutoff bands. Compare seeds within panels; r8/r10 comparisons are historical and tasks differ. Paired probe intervals: Appendix E.}
\expandafter\def\csname captionfig4\endcsname{Target noise, access and encoding affect measured errors without guaranteeing prediction closure. Matched r13 comparisons: targets within versions (a,b), numbered access variants (c–e), encodings of exact sums (f). Shapes: seeds 0/1/2; grey crosses: matched update 0; dashed line: zero-error oracle except in (d), whose 0-bit oracle lies outside the logarithmic axis. Categorical misreads: 24,435 boards, rendering 0; natural KL: 60,000 positions; maximum long-panel TV: 384 N3–N8 endpoints, bar 0.02. No pooled means or bands. Controls are matched within panels.}
\expandafter\def\csname captionfig5\endcsname{Prediction gradients erode exact sums at update 1, not uniquely under AdamW. Columns: seeds 0–2. (a–c) Compare r11 losses: exact starts on 24,435 boards; updates 1/5/10/50/100 use a fixed 128-board subset. Symlog separates 0/1. (d–f) r13, all 1,555 local histories: every tested active optimizer loses exactness at update 1. Update-1 correct counts, seeds 0/1/2, each out of 1,555: AdamW: 393 / 5 / 518; reduced-lr AdamW: 1,554 / 1,468 / 1,553; SGD: 1,440 / 89 / 1,495; blocked prediction gradient: 1,555 / 1,555 / 1,555. Blocked prediction gradient is a separate weight-decay control; dashed line: exact start/oracle. Movement-matched points: \cref{fig:appendix_D_movement}. r11/r13 is historical; no pooling or bands.}
\expandafter\def\csname captionappendix_D_movement\endcsname{Exact sums erode across active optimizers at common recorded movement. (a–c) Seeds 0–2, all 1,555 local histories. Compare AdamW, reduced-rate AdamW and SGD within each seed's three-active-arm common range; points are observed, not interpolated. Blocked prediction gradient is excluded from that range and remains a separate control in \cref{fig:fig5}. Dashed line: exact start/oracle. No pooled means or confidence bands.}
\expandafter\def\csname captionappendix_C_methods\endcsname{Separate probe (a) and swap-test (b) procedures. Readers and alignment fit on training identities only; scoring identities are held out. Positive, no-op and shuffled controls calibrate the procedure, while donor-follow and selectivity concern the evaluated carrier. This schematic has no empirical seed average or accuracy; full controls and support are in Appendix C.}
\expandafter\def\csname captionappendix_E_geometry\endcsname{Cutoff discrimination can succeed while prediction still fails. The supplementary panel was bound before states were read, with disjoint fit/evaluation identities and fixed renderings. Three seed columns show saved 0/5k/20k sampled- and probability-target prediction-loss checkpoints only, with no network updates: (a–c) accuracy and (d–f) cross-entropy, in seed order 0–2. Compare sampled- and probability-target readers for the same seed and pair; circles/solid lines are 12/13 (135 fitting pairs, 150 evaluation pairs; 300 endpoint labels), squares/dashed lines are 23/24 (12/13 pairs; 26 labels). Grey crosses are the saved shuffled-label floors; balanced constant floors are 0.5 accuracy and ln 2 loss, and the exact ceiling is 1 accuracy / 0 loss. No confidence bands are implied. Distances and rendering variation remain descriptive in Appendix E's support tables; successful discrimination is not proof of predictive use or of a precision ceiling.}
\expandafter\def\csname captionappendix_D_r8\endcsname{Recorded r8 learning curves show changes in average error, not sufficient evidence of reliable competence. Every seed and condition is retained; (a–c) original and (d–f) equal laws are separate rows, in seed order 0–2. Open points mark each condition's own update-0 control; dashed zero is the exact oracle. The quantity is natural KL in bits on 5,000 natural episodes of 12 rounds (60,000 scored positions), not a loss or a census frequency. No smoothing, selected checkpoint or across-seed band is used. Compare conditions within the same seed and law only; cross-study comparisons are historical. Natural KL alone does not establish reliability.}
\expandafter\def\csname captionappendix_D_r10\endcsname{Recorded r10 learning curves show changes in average error, not sufficient evidence of reliable competence. Every seed and condition is retained; (a–c) original and (d–f) equal laws are separate rows, in seed order 0–2. Open points mark each condition's own update-0 control; dashed zero is the exact oracle. The quantity is natural KL in bits on 5,000 natural episodes of 12 rounds (60,000 scored positions), not a loss or a census frequency. No smoothing, selected checkpoint or across-seed band is used. Compare conditions within the same seed and law only; cross-study comparisons are historical. Natural KL alone does not establish reliability.}
\expandafter\def\csname captionappendix_D_r11_rarity\endcsname{Recorded r11 rarity learning curves show changes in average error, not sufficient evidence of reliable competence. Every seed and condition is retained; (a–c) original and (d–f) equal laws are separate rows, in seed order 0–2. Open points mark each condition's own update-0 control; dashed zero is the exact oracle. The quantity is natural KL in bits on 5,000 natural episodes of 12 rounds (60,000 scored positions), not a loss or a census frequency. No smoothing, selected checkpoint or across-seed band is used. Compare conditions within the same seed and law only; cross-study comparisons are historical. Natural KL alone does not establish reliability.}
\expandafter\def\csname captionappendix_D_r11_dose\endcsname{Recorded r11 dose learning curves show changes in average error, not sufficient evidence of reliable competence. Every seed and condition is retained; (a–c) original and (d–f) equal laws are separate rows, in seed order 0–2. Open points mark each condition's own update-0 control; dashed zero is the exact oracle. The quantity is natural KL in bits on 5,000 natural episodes of 12 rounds (60,000 scored positions), not a loss or a census frequency. No smoothing, selected checkpoint or across-seed band is used. Compare conditions within the same seed and law only; cross-study comparisons are historical. Natural KL alone does not establish reliability.}
\expandafter\def\csname captionappendix_D_r11_erosion\endcsname{Recorded r11 erosion learning curves show changes in average error, not sufficient evidence of reliable competence. Every seed and condition is retained; (a–c) original and (d–f) equal laws are separate rows, in seed order 0–2. Open points mark each condition's own update-0 control; dashed zero is the exact oracle. The quantity is natural KL in bits on 5,000 natural episodes of 12 rounds (60,000 scored positions), not a loss or a census frequency. No smoothing, selected checkpoint or across-seed band is used. Compare conditions within the same seed and law only; cross-study comparisons are historical. Natural KL alone does not establish reliability.}
\expandafter\def\csname captionappendix_D_r12\endcsname{Recorded r12 learning curves show changes in average error, not sufficient evidence of reliable competence. Every seed and condition is retained; (a–c) original and (d–f) equal laws are separate rows, in seed order 0–2. Open points mark each condition's own update-0 control; dashed zero is the exact oracle. The quantity is natural KL in bits on 5,000 natural episodes of 12 rounds (60,000 scored positions), not a loss or a census frequency. No smoothing, selected checkpoint or across-seed band is used. Compare conditions within the same seed and law only; cross-study comparisons are historical. Natural KL alone does not establish reliability.}
\expandafter\def\csname captionappendix_D_r13_noise\endcsname{Recorded r13 noise learning curves show changes in average error, not sufficient evidence of reliable competence. Every seed and condition is retained; (a–c) original; no equal-law arms were run, in seed order 0–2. Open points mark each condition's own update-0 control; dashed zero is the exact oracle. The quantity is natural KL in bits on 5,000 natural episodes of 12 rounds (60,000 scored positions), not a loss or a census frequency. No smoothing, selected checkpoint or across-seed band is used. Compare conditions within the same seed and law only; cross-study comparisons are historical. Natural KL alone does not establish reliability.}
\expandafter\def\csname captionappendix_D_r13_access\endcsname{Recorded r13 access learning curves show changes in average error, not sufficient evidence of reliable competence. Every seed and condition is retained; (a–c) original; no equal-law arms were run, in seed order 0–2. Open points mark each condition's own update-0 control; dashed zero is the exact oracle. The quantity is natural KL in bits on 5,000 natural episodes of 12 rounds (60,000 scored positions), not a loss or a census frequency. No smoothing, selected checkpoint or across-seed band is used. Compare conditions within the same seed and law only; cross-study comparisons are historical. Natural KL alone does not establish reliability.}
\expandafter\def\csname captionappendix_D_r13_encoding\endcsname{Recorded r13 encoding learning curves show changes in average error, not sufficient evidence of reliable competence. Every seed and condition is retained; (a–c) original and (d–f) equal laws are separate rows, in seed order 0–2. Open points mark each condition's own update-0 control; dashed zero is the exact oracle. The quantity is natural KL in bits on 5,000 natural episodes of 12 rounds (60,000 scored positions), not a loss or a census frequency. No smoothing, selected checkpoint or across-seed band is used. Compare conditions within the same seed and law only; cross-study comparisons are historical. Natural KL alone does not establish reliability.}

\hypersetup{hidelinks,pdfauthor={Ioannis Tsiokos},
  pdftitle={Where Does Jagged Competence Come From?}}
\fancypagestyle{firstpage}{\fancyhf{}
  \fancyfoot[C]{\scriptsize Automorph Inc., Wilmington, DE, USA\quad
  \texttt{ioannis@automorph.io}\quad\textcopyright\ \manuscriptyear\ $\cdot$\ CC-BY 4.0}
  }
\title{Where Does Jagged Competence Come From?}
\author{Ioannis Tsiokos\,\orcidlink{0009-0009-7659-5964}}
\date{\manuscriptdate\quad\manuscriptversion\ifpaperhasdoi\quad DOI: \href{https://doi.org/\paperdoi}{\paperdoi}\fi}
\begin{document}
\maketitle
\thispagestyle{firstpage}
\begin{abstract}
Capable systems often show jagged competence: low average error alongside
failures on particular inputs. We ask where it comes from in a task built from
two known layers. A lower layer A computes five per-slot sums from records;
an upper layer B uses the slot-1 sum and a mode carried over from earlier
boards to predict the next board's category, so B cannot be computed from the
current A alone: B is a strict extension of A. We train small recurrent
networks on B and measure, against exact ground truth, what they acquire of
A. The central finding is that learning B gives the network a jagged version
of A, measured through probe readability and supervised outputs. It is
readable where B needs it (slot 1, 97--98\% by a linear probe) and becomes
less readable where B does not (the other slots, 5--10\%); category misreads
concentrate near the slot-1 cutoffs; and when A is trained explicitly it
comes out only approximately right. The networks show jagged competence in
B: natural KL below $3\times10^{-4}$ bits coexists with a maximum
law TV of about 0.25 on constructed histories. Matched experiments show that
even a good A is not enough: connecting a learned A to B cuts misreads
1.3--17-fold, the same exact A gives fewer misreads as one-hot inputs (0--2)
than as numerical inputs (10--317), one B update makes an exact A inexact unless the B
gradient is blocked, and exact A still leaves some B failures. In this task,
uneven acquisition, access and preservation of the lower theory explain part
of the jagged competence of a network that learns the theory built on it;
some failures remain unexplained. Whether the same
holds in larger systems is a test to run.
\end{abstract}
\section{Introduction}\label{sec:intro}

Jagged competence combines strong performance with failures on particular
inputs.\footnote{\citet{gans2026} credits the term ``jagged intelligence''
to Andrej Karpathy; we use ``competence'' for performance in this task.}
Field experiments document an uneven AI frontier \citep{dellacqua2023}, and
\citet{gans2026} models how average quality can hide local unreliability. The
pattern is easy to observe and hard to explain: in a large system we rarely
know which intermediate quantities the task requires, or whether the system
computes, uses and keeps them, and without that knowledge ``jagged'' remains
a description.

We study the question in a task built from two known layers. Layer A computes
five per-slot sums from a board of records. Layer B uses the slot-1 sum's
category and a mode carried over from earlier boards to set the probabilities
of the next board's category. The same board can need different predictions
after different histories, so B cannot be computed from the current A alone;
in the terms of Six Birds Theory, B is a strict extension of A
(\cref{sec:strict}). A working hypothesis of that programme, motivated by its
picture of theory levels \citep[\S1.1]{tsiokos2026foundations} rather than
proved for trained networks, is that a network trained on an upper theory
comes to know the theory beneath it. Because both layers are known exactly,
we can test this: train networks on B only, and measure what they acquire of
A and how it relates to where B fails.

The answer, in this task, is that learning B gives the network a jagged
version of A, and that this explains part of the jagged competence in B. The
network makes readable the part of A that B needs while the rest becomes
less readable; in r12's uniform-sampling census, category misreads
concentrate near the slot-1 cutoffs; and when A is trained explicitly, it
comes out largely but not exactly right. The networks reach a natural KL below
$3\times10^{-4}$ bits, more than thirty times below our 0.01-bit bar, yet
fail on constructed histories (maximum law TV 0.25--0.26), and their category
misreads sit near the slot-1 cutoffs (\cref{sec:res-recover,sec:res-average}).
Here ``acquire'' means what we measure, readable by probes and approximately
computed when supervised; we do not demonstrate an internal layer, and the
jagged A explains part of B's failures, not all of them.

A second lesson is that knowing an intermediate quantity is not one thing.
A network can hold A in readable form without established exact computation,
compute it without delivering it to the prediction, deliver it in a form that
supports less reliable prediction, or lose it under further training; and
even an exact, delivered A leaves some failures in B. These are separate
achievements, not a required sequence, and matched comparisons isolate the
effects of connection, encoding and prediction gradients
(\cref{sec:res-noise,sec:res-access,sec:res-encoding,sec:res-erosion,sec:res-reliable}).

\paragraph{Contributions.}
\begin{itemize}
\item The finding that learning a strict extension gives a network only a
  jagged version of the lower theory, and that this explains part of
  the upper one's failure, with the unexplained cases stated
  (\cref{sec:results,sec:discussion}).
\item A decomposition of what it means for a network to know an intermediate
  quantity into five measurable achievements (recoverability, computation,
  access, preservation and reliable prediction), with matched experiments on
  access, encoding, preservation and target noise (\cref{sec:results}).
\item A layered task with exhaustive board ground truth and calibrated
  measurements (\cref{sec:task,sec:measurements}), and a provided ledger and
  code rebuilding every reported number, figure and table (\cref{app:compute}).
\end{itemize}

\section{Related work}\label{sec:related}

\paragraph{Jagged competence.} \citet{dellacqua2023} document an uneven AI
frontier in a field experiment, and \citet{gans2026} models uneven local
reliability through gaps in knowledge coverage. \citet{kalai2024} derive
calibration-related bounds on errors about arbitrary facts, while
distinguishing systematic facts. Our task has a systematic rule, so those
bounds do not explain its errors.

\paragraph{What training makes available.} The training task shapes which
input features a network represents and how readable they are
\citep{hermann2020}, and gradient descent can favor simple features and
neglect others \citep{shah2020,pezeshki2021,geirhos2020}. We ask which parts of a designed intermediate computation become
recoverable under final-output supervision, and whether recoverability
accompanies computation, access and preservation.

\paragraph{Probing versus computation and use.} Probe accuracy alone does
not show that a network uses what the probe finds: probes need controls for
their own capacity \citep{hewitt2019,belinkov2022}, high probing accuracy can
coexist with task irrelevance \citep{ravichander2021}, and removing
information tests use more directly \citep{elazar2021}. \citet{darade2026}
recover candidate arithmetic intermediates but do not find them transmitted
along the tested causal routes, and \citet{basu2026} report a dissociation
between probe discrimination and output behavior. Interchange interventions test whether an aligned internal variable plays a
causal role \citep{geiger2021,geiger2024}, and causal tracing and model
editing locate and modify stored associations \citep{meng2022}; our swap test follows
this design but failed selectivity under the tested alignment, a limit of
the test rather than evidence of absence. We separately measure exact output
computation and predictive access, then test whether connecting a learned
sums output reduces errors.

\paragraph{Controlled tasks with known ground truth.} Controlled tasks allow
exact comparison of internal states and behavior with a reference:
board-game world states \citep{li2023}, the gap between next-token accuracy
and a coherent world model \citep{vafa2024}, extrapolation on formal
languages \citep{deletang2023}, synthetic reasoning \citep{allenzhu2025},
grokking and superposition \citep{nanda2023,elhage2022}. \citet{shintani2026}
separates arithmetic failures into stages tested with matched data
interventions. We share this controlled-decomposition approach, while
studying a recurrent layered task with measures of recoverability, output
computation, access and preservation.

\paragraph{Auxiliary targets, soft targets and interference.} Soft targets
carry more information than sampled labels \citep{hinton2015}, objectives can
conflict \citep{yu2020}, and training on a new objective can degrade an
earlier one \citep{mccloskey1989,kirkpatrick2017}. \citet{shu2026} propose an
account of jaggedness involving uneven optimization allocation and
capability coupling. Our erosion experiment establishes dependence on the
prediction gradient, but does not identify interference as its mechanism.

\section{Task and setup}\label{sec:task}
\begin{figure}[tb]
\centering
\includegraphics[width=\textwidth]{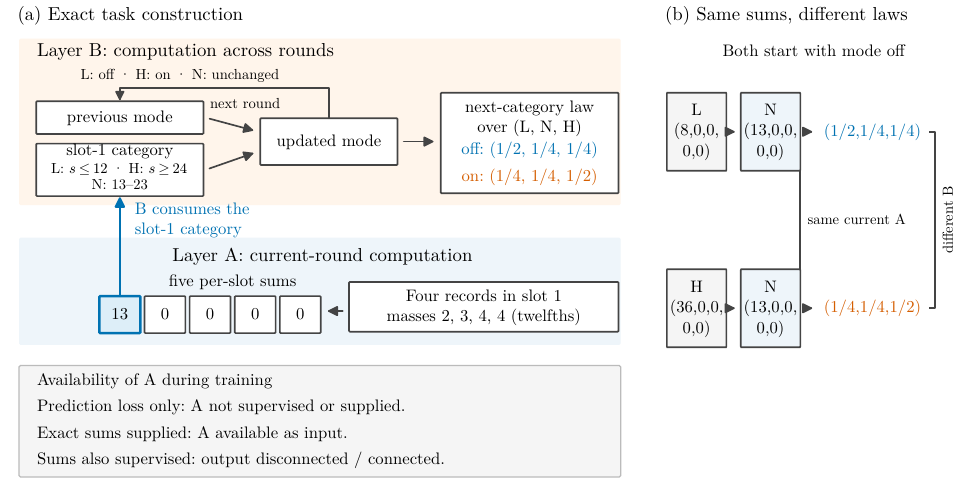}
\caption{\csname captionfig1\endcsname}\label{fig:fig1}
\end{figure}

\begin{figure}[tb]
\centering
\includegraphics[width=\textwidth]{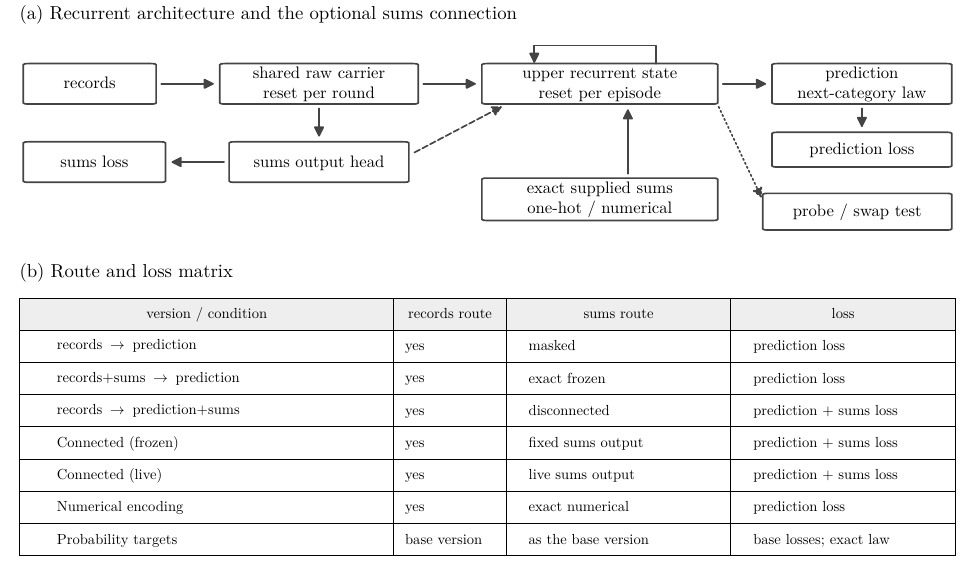}
\caption{\csname captionfig2\endcsname}\label{fig:fig2}
\end{figure}

In the records game, the quantities a network might compute, the hidden
variable and the correct next-step probabilities are known exactly for every
input, and all boards can be enumerated (\cref{fig:fig1}).

\subsection{The game}\label{sec:game}

\paragraph{Boards and sums.} A \emph{record} places one mass in one of five
\emph{slots}. The six masses are $1/6$, $1/4$, $1/3$, $1/2$, $2/3$ and $3/4$,
written below in integer twelfths (2, 3, 4, 6, 8, 9). A \emph{board} is an
ordered sequence of four records. Its \emph{sums} are the five per-slot
totals, each one of the 36 attainable values $0$ and $2$--$36$ twelfths.
Computing the sums defines layer~A. The $30^4 = 810{,}000$ ordered placements
yield 24,435 distinct \emph{completed sum boards}; many record sequences
realize the same completed sum board, and we call one such realization a
\emph{rendering}.

\paragraph{Category, mode and law.} Only slot~1 matters for prediction. Its
sum gives the board's \emph{slot-1 category}: \Lcat{} when the sum is at most
12 twelfths, \Hcat{} when it is at least 24, and \Ncat{} otherwise (13--23).
The task keeps a binary \emph{mode}, which starts off. An \Hcat{} board
switches it on, an \Lcat{} board switches it off, and an \Ncat{} board leaves
it unchanged. The next board's category is drawn from the \emph{law}
$(\tfrac12, \tfrac14, \tfrac14)$ over $(\Lcat, \Ncat, \Hcat)$ when the mode
is off and $(\tfrac14, \tfrac14, \tfrac12)$ when it is on. Given the
category, the next board is a uniformly chosen completed sum board of that
category, rendered as a legal placement of records. At every round the
network sees the current board's records and outputs a \emph{prediction}: a
probability vector over the next board's category. Predicting the law is the
upper task layer~B. In the \emph{equal-law control} both modes use
$(\tfrac38, \tfrac14, \tfrac38)$, so the mode is irrelevant; it checks that
the measurements do not report mode use where none is possible.

\paragraph{Boards near the cutoffs.} 22,491 completed sum boards are
\Lcat{}, 1,835 are \Ncat{} and 109 are \Hcat{}. The \emph{cutoff band}
(slot-1 sums 11--13 and 23--25; 930 boards) is 2.5\% of the \Lcat{} boards,
16.9\% of the \Ncat{} boards and 45.9\% of the \Hcat{} boards. Under uniform
sampling within a category, these shares are also the training probability
of a band board; the rarity conditions of r11 and r12 change them.

\subsection{Two task layers}\label{sec:strict}

Layer A maps each round's records to its five sums. Layer B operates over
sequences of these completed sum boards: the slot-1 category updates the
preceding mode, and the updated mode determines the next-category law
(\cref{fig:fig1}a). The exact update needs only two things, the current
board's slot-1 category and the previous mode. A network that predicts
reliably must therefore combine information about the current board's
category with information from earlier boards; it need not do so through the
exact sums.

Six Birds Theory gives this relationship a precise meaning. A description B
\emph{strictly extends} a description A when B draws distinctions that no
function of A can draw. The theory uses this as its certificate of novelty:
a new level of description adds predicates that are not definable from the
old one, rather than computing further within it
\citep[\S1.1]{tsiokos2026foundations}. Formally, B does not factor through A,
which holds exactly when some pair of cases has equal A and different B
(\citealp[Theorem~12]{tsiokos2026foundationsiii};
\citealp[\S3.5]{tsiokos2026strict}); this non-factorization is the only
condition we use. Under the original law, B (the next-category law) strictly
extends the current A description (the five sums): no function of the
current sums gives the correct law for every history. Two histories ending on
the same neutral board have identical sums but can retain different modes
and require different laws (\cref{fig:fig1}b). B is determined by the
sequence of A objects and the initial mode, not by the current A object
alone. Neither containment nor recovery of A from B is required. The
equal-law control removes this distinction.

These are the task's layers. Whether training realizes their decomposition
internally is the empirical question.

\subsection{Network, versions and training}\label{sec:versions}

\paragraph{Network.} For the original task all networks share one
architecture (\cref{fig:fig2}); in r10 the upper state also receives the
public query. Within a round, a one-layer GRU of width 16 reads the four
records and produces the \emph{raw carrier}, which is reset at every round.
Across rounds, an upper GRU cell of width 32 receives the raw carrier and,
where a version provides them, the five sums as one-hot vectors; a linear
head on the upper state gives the prediction. Supplied sums come from a
separately trained sums network that is exact on all 1,555 local histories
(every ordered sequence of up to four masses in one slot) and is kept
frozen. Details are in \cref{app:training}.

\paragraph{Versions.} Each version name states the input, an arrow, and what
the loss is computed on. In \textbf{records $\to$ prediction} the input is
the records and the loss is on the prediction only. In \textbf{records+sums
$\to$ prediction} the input also includes the exact one-hot sums. In
\textbf{records $\to$ prediction+sums} the input is the records and the loss
is on the prediction plus a \emph{sums output}, a linear head on the raw
carrier trained to give the five sums and not fed to the prediction.
r8 trained the first two versions and a third, \textbf{records+sums $\to$
prediction (sums route only)}, whose upper state receives only the exact
sums; records $\to$ prediction+sums first appears in r10. The versions vary the supervision of A and
its availability to B: no explicit sums supervision, exact supplied sums, or
supervised sums outputs disconnected from or connected to prediction.
Further variants change one factor. \textbf{Connected (frozen)} and \textbf{Connected (live)}
feed the sums output of a trained records $\to$ prediction+sums network into
the prediction, from a fixed copy or from the sums output being trained.
\textbf{Numerical encoding} gives the exact sums as $(s/36,\,1-s/36)$ per
slot instead of one-hot. \textbf{Probability targets} train against the
exact law instead of one sampled next category. The \textbf{erosion
continuation} starts from an exact sums network that is no longer frozen and
updates either the prediction loss only or the prediction plus sums loss.

\paragraph{Training.} Full training runs use three seeds (0, 1, 2), a fixed
20,000 updates, AdamW (learning rate 0.003, weight decay 0.01, batch 256),
float32 and freshly drawn episodes; no checkpoint is selected by accuracy.
From r12 on, each batch also contains \emph{history coverage}: constructed
prefixes of an \Lcat{} or \Hcat{} board followed by up to eight \Ncat{}
boards. The r13 erosion replays instead run for 200 updates and compare
three optimizers for the sums network (AdamW, AdamW with a tenfold smaller
learning rate, and SGD matched to the norm of AdamW's first gradient-only displacement) with a control that
blocks the prediction gradient to the sums network while weight decay
continues. r9 trains nothing. Details are in \cref{app:training}.

\subsection{Study history and comparisons}\label{sec:history}

The evidence comes from six studies, r8--r13, each registered with its
endpoints and criteria before it ran. r8 trained the three r8 networks under
both laws. r9 diagnosed the saved r8 networks. r10 retains A but changes B to the
\emph{all-slots task}, testing readability and use of all five sums: there
each slot remembers its last nonzero sum (initially zero) and the law over
the next board's total sum modulo three depends on the remembered value of a
publicly queried slot, not its current sum (36 distinct laws). r11 varied board
rarity near the cutoffs, the amount of sums supervision, and the erosion
continuation. r12 added history coverage. r13 ran four mechanism experiments:
target noise, access, encoding and erosion. Follow-ups, corrected audits and
supplementary analyses are labeled in \cref{app:history}; the campaign was
not one confirmatory study.

A comparison is \emph{matched} when the code, data streams, initialization,
number of updates and evaluation panels are identical and only the stated
factor differs. All other comparisons are \emph{historical}. Comparisons
between r11 and r12 change the curriculum and the loss weighting at once;
comparisons between r10 and the other studies change the task. The r13
access and encoding experiments share one reference condition, the exact
one-hot supplied sums, which we count once.

\section{Measurements}\label{sec:measurements}

Each reported effect is read against a floor and a ceiling. For prediction,
the ceiling is the exact law (zero KL and zero law error) and the main floor
is the \emph{board-only predictor}, which knows the current board exactly
but not the history. For probes, the floors are the majority class,
shuffled labels and the network at update~0, and the ceiling is the same
probe fitted on an exact sums carrier.

\paragraph{Prediction: average and constructed tests.} \emph{Natural KL} is
$\KL(\text{law}\,\|\,\text{prediction})$ in bits, averaged over 60,000
positions (5,000 test episodes of 12 rounds); \emph{unseen KL} is the same
quantity on test positions whose preceding pair of boards never occurred in
training. Average error can be low while specific histories fail, so we also
score \emph{law panels}: constructed histories made of an \Lcat{} or \Hcat{}
anchor board followed by $k$ \Ncat{} boards, $k = 0,\dots,8$, with 32
variants per anchor and length (576 histories). The \emph{law TV} is the
total-variation distance between the prediction and the law at the end of a
history. The \emph{short panel} covers the anchors and
$\Ncat^1$--$\Ncat^2$; the \emph{long panel} covers $\Ncat^3$--$\Ncat^8$.

\paragraph{Passing.} A run \emph{passes} when it meets every prediction
criterion registered for its study at the fixed 20,000-update endpoint. For
the original task under the original law these are: natural and unseen KL at
most 0.01 bits; maximum law TV at most 0.02; witness recovery (the share of
the history effect captured one step after the anchor) at least 0.8; and
prediction TV at most 0.02 under rerendered boards and on the registered
state-intervention panels. The equal-law control and r10 use their own
criteria; r9 and the erosion replays have no pass endpoint. Complete
criteria for every study are in the supplement (\cref{app:measurement}). A pass is a statement about these criteria, not
exhaustive correctness.

\paragraph{Census.} To locate failures on boards rather than histories, we
score every completed sum board after an \Lcat{} anchor and after an \Hcat{}
anchor. The two predictions form the board's \emph{signature}. A
\emph{categorical misread} is a board whose signature is closer to the exact
signature of a wrong category than to that of its own. A \emph{TV-error
count} is a board whose larger prediction TV over the two contexts exceeds
0.02. These are different measures and we never merge them: a board can miss
the 0.02 bar without being misread. Misread signatures are undefined under
the equal law.

\paragraph{Recoverability.} A \emph{probe} is a classifier fitted after
training on frozen internal states, the raw carrier or the upper state, that
predicts one slot's sum. It is fitted on 2,048 boards and scored on 512
other boards per slot. We use linear probes and one-hidden-layer probes of
width 64. A sum is \emph{recoverable} when the probe scores above the
majority-class and shuffled-label floors. Separately, the probe's
\emph{gain} is its improvement over the same probe on the network at
update~0, with a paired 95\% interval over the evaluation boards. That
interval describes probe uncertainty for one network, not variation across
seeds, which we report as per-seed values.

\paragraph{Computation and preservation.} \emph{Local exactness} is the
fraction of the 1,555 local histories for which a sums output is exactly
correct. \emph{Board-vector exactness} is the fraction of the 24,435
completed sum boards whose five-sum vector is correct; the early readouts of
the r11 erosion continuation use a fixed sample of 128 boards instead.
Preservation is the change in these quantities during further training from
an exact start. To compare optimizers whose steps differ in size, we track
\emph{cumulative movement}: the sum over updates of the Euclidean norm of
each parameter step of the sums network.

\paragraph{Access and use.} Recoverable is not used. We say the prediction
\emph{uses} a quantity only when one of two kinds of evidence supports it: a
matched comparison that changes only how the quantity reaches the prediction
(connected versus disconnected, one-hot versus numerical), or a calibrated,
selective swap test. In the swap test (r10), the raw carrier of the
networks that read only records is mapped by a fitted orthogonal alignment onto
coordinates that track the five sums, and the three-coordinate block for the
queried slot is replaced with the block
from a donor board that differs only in that slot's value. \emph{Donor-follow} is the fraction
of accurate pairs for which the patched prediction is closer to the donor's
law than to the base board's; \emph{selectivity} requires that the other
four slots' decoded sums do not change. Calibration on a trained positive
and two negatives, selectivity, and the evaluated result are reported
separately (\cref{app:measurement,app:calibration}). A swap that fails
selectivity under one alignment does not show that the quantity is absent or
unused; it shows only that this test cannot support use.

\paragraph{Calibration.} Every bar was checked by running its deciding
function on known positive and null cases before results were interpreted.
A calibration pass shows that an instrument can discriminate; it is not a
result about any network.

\section{Results}\label{sec:results}
\begin{table}[tb]
\centering
\caption{Five achievements, original law. $\checkmark$ supported; $\times$ not supported; – not established: available evidence does not establish this achievement. Fractions are seeds passing out of three, except recoverability: supported slots out of five in seed order 0/1/2. Access: $D$ by design (exact sums supplied); $I$ demonstrated by a matched connection intervention. Complete scope, counts and controls: Supplement Table S1.}\label{tab:table1}
\begingroup\fontsize{8}{9.5}\selectfont\setlength{\tabcolsep}{3pt}\emergencystretch=1em\hyphenpenalty=10000\exhyphenpenalty=10000
\begin{tabular}{@{}>{\raggedright\arraybackslash}p{2.500in}>{\raggedright\arraybackslash}p{0.910in}>{\raggedright\arraybackslash}p{0.780in}>{\raggedright\arraybackslash}p{0.430in}>{\raggedright\arraybackslash}p{0.720in}>{\raggedright\arraybackslash}p{0.740in}@{}}
\toprule
\textbf{condition} & \textbf{recoverability} & \textbf{computation} & \textbf{access} & \textbf{preservation} & \textbf{reliable prediction} \\
\midrule
r8: records $\to$ prediction & $\checkmark$ 1/5$\cdot$1/5$\cdot$1/5 & – & – & – & $\times$ 0/3 \\
r10: records $\to$ prediction & $\checkmark$ 5/5$\cdot$5/5$\cdot$5/5 & – & – & – & $\times$ 0/3 \\
r10: records $\to$ prediction+sums & $\checkmark$ 5/5$\cdot$5/5$\cdot$5/5 & – & – & – & $\times$ 0/3 \\
r11: records $\to$ prediction (Erosion continuation) & $\checkmark$ 1/5$\cdot$1/5$\cdot$1/5 & $\times$ 0/3 & – & $\times$ & $\times$ 0/3 \\
r11: records $\to$ prediction+sums (Erosion continuation) & $\checkmark$ 3/5$\cdot$1/5$\cdot$1/5 & $\checkmark$ 1/3 & – & $\times$ & $\checkmark$ 1/3 \\
r12: records $\to$ prediction+sums & $\checkmark$ 5/5$\cdot$5/5$\cdot$5/5 & $\times$ 0/3 & – & – & $\times$ 0/3 \\
r12: records+sums $\to$ prediction & $\checkmark$ 2/5$\cdot$2/5$\cdot$3/5 & $\checkmark$ 3/3 & $\checkmark^{D}$ & $\checkmark^{f}$ & $\checkmark$ 2/3 \\
r13: records $\to$ prediction+sums & – & $\times$ 0/3 & – & – & $\times$ 0/3 \\
r13: Connected (frozen) & – & $\times$ 0/3 & $\checkmark^{I}$ & $\checkmark^{f}$ & $\times$ 0/3 \\
r13: Connected (live) & – & $\times$ 0/3 & $\checkmark^{I}$ & – & $\times$ 0/3 \\
r13: records+sums $\to$ prediction (exact supplied) & – & $\checkmark$ 3/3 & $\checkmark^{D}$ & $\checkmark^{f}$ & $\checkmark$ 1/3 \\
r13: records+sums $\to$ prediction (Numerical encoding) & – & $\checkmark$ 3/3 & $\checkmark^{D}$ & $\checkmark^{f}$ & $\times$ 0/3 \\
r13: blocked prediction gradient & – & $\checkmark$ $3/3^{\ell}$ & – & $\checkmark^{b}$ & – \\
\bottomrule
\end{tabular}
\par\smallskip
\begin{minipage}{\textwidth}\textit{Note.} Raw-carrier linear readers: 512 held-out boards/slot. Computation: the audited sums output, exact vectors on rendering 0 (24,435 boards); $\ell$: 1,555 local histories. The connection intervention does not establish selective use. $f$: frozen supplier; $b$: blocked-gradient control. Frozen preservation is design-limited.\end{minipage}
\endgroup
\end{table}

\Cref{tab:table1} summarizes, for each version and condition, which of the
five achievements of realizing the designed decomposition the evidence
supports; its columns follow the order used throughout (recoverability, computation, access,
preservation, reliable prediction). Unless stated otherwise, results
use the original law, the 20,000-update endpoint and census rendering~0,
and values are per seed in seed order (0 / 1 / 2).

\subsection{Training shapes what becomes recoverable}\label{sec:res-recover}
\begin{figure}[tb]
\centering
\includegraphics[width=\textwidth]{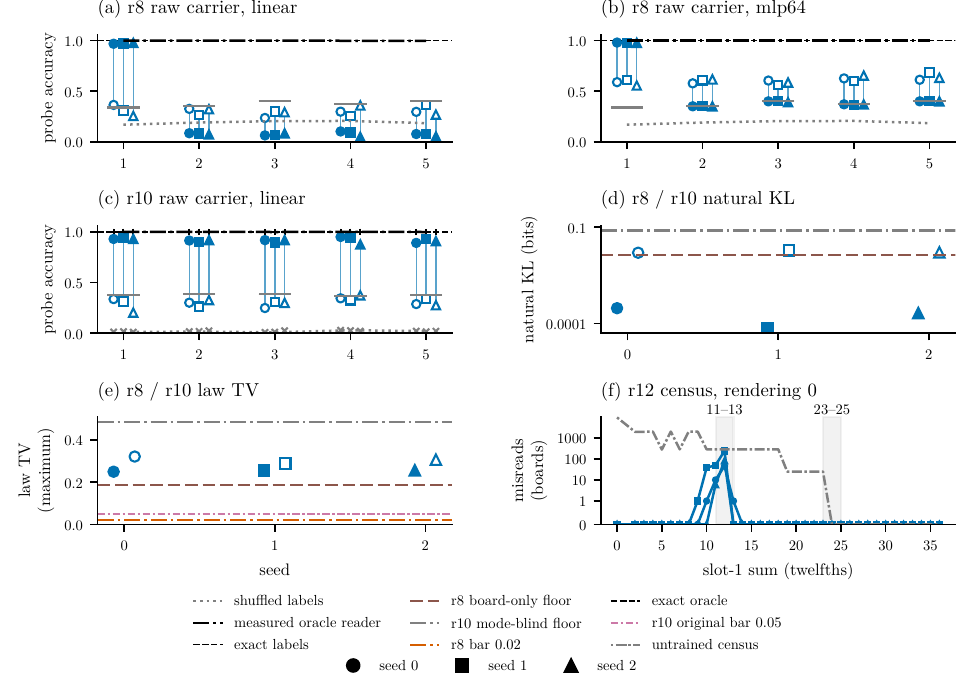}
\caption{\csname captionfig3\endcsname}\label{fig:fig3}
\end{figure}

\textbf{Claim: training on B gives the network a jagged version of A,
readable where B needs it and less readable where it does not.} In the
records $\to$ prediction networks of r8, a linear probe on the raw carrier
recovers the slot-1 sum on 97--98\% of held-out boards at the endpoint, up
from 26--37\% at update~0 (ranges over three seeds; \cref{fig:fig3}a). The sums of slots 2--5, which
the prediction never needs, become \emph{less} readable: from 24--37\% at
update~0 (mean 30\%) to 5--10\% at the endpoint (ranges over four slots
and three seeds). A one-hidden-layer probe shows the same
direction (about 60\% to 38\%, mean over slots 2--5 and three seeds;
\cref{fig:fig3}b). The linear probe weights classes equally, so part of the
drop below the majority floor reflects that weighting; we read the decline as
reduced measured readability, not as removal of information. In the
all-slots task (r10), where the law depends on whichever slot is queried,
linear probes on the raw carrier recover all five sums: 20--37\% at
update~0 and 88--95\% at the endpoint over five slots and three seeds, with
every paired 95\% gain interval above zero (\cref{fig:fig3}c).

\emph{Scope.} The r8--r10 comparison is historical and crosses tasks. It
supports task-dependent organization of what is recoverable, not exact
computation or use; rare r10 sum values (28--36) lie outside the registered
scope.

\subsection{Average error hides specific failures}\label{sec:res-average}

\textbf{Claim: B combines low average error with category misreads
concentrated near the slot-1 cutoffs and failures on particular histories.}
The r8 records $\to$ prediction networks reach a natural KL of
0.00030 / 0.00007 / 0.00021 bits, far below the 0.01 bar and the board-only
predictor's 0.0135 bits (\cref{fig:fig3}d). Yet none of the three runs
passes: on the constructed law panels the maximum law TV is
0.250 / 0.255 / 0.256, worse than the board-only predictor (0.188) and
twelve times the 0.02 bar (\cref{fig:fig3}e). A separate census of the r12 records $\to$ prediction networks (uniform
sampling) locates the board failures: 57 of 58, 289 of 328 and 87 of 87
categorical misreads lie in the cutoff band; the rest lie at slot-1 sums 9
and 10, just below it (\cref{fig:fig3}f).

One might expect training on long histories to remove the history
failures. r12 added history coverage, and its long-panel errors are smaller
than r11's (a historical comparison: curriculum and loss weighting also
changed). Even so, none of the nine
r12 records $\to$ prediction runs passes (three board-sampling arms $\times$
three seeds).
Long-panel maxima reach 0.172 and short-panel maxima range 0.020--0.066. We
diagnosed 162 selected histories from these runs. 120 stay within the
diagnostic 0.02-TV bar. Of the remaining 42, 20 are board-associated (all
coincident with a misread of the current board), 3 are
recurrence-associated, and 19 are unresolved. Misreads therefore do not
account for every failure.
Enriching training with boards at the cutoff values (r11) did not repair the
misreads consistently across seeds (\cref{app:tables}).

\subsection{Target noise contributes but is not sufficient}\label{sec:res-noise}
\begin{figure}[tb]
\centering
\includegraphics[width=\textwidth]{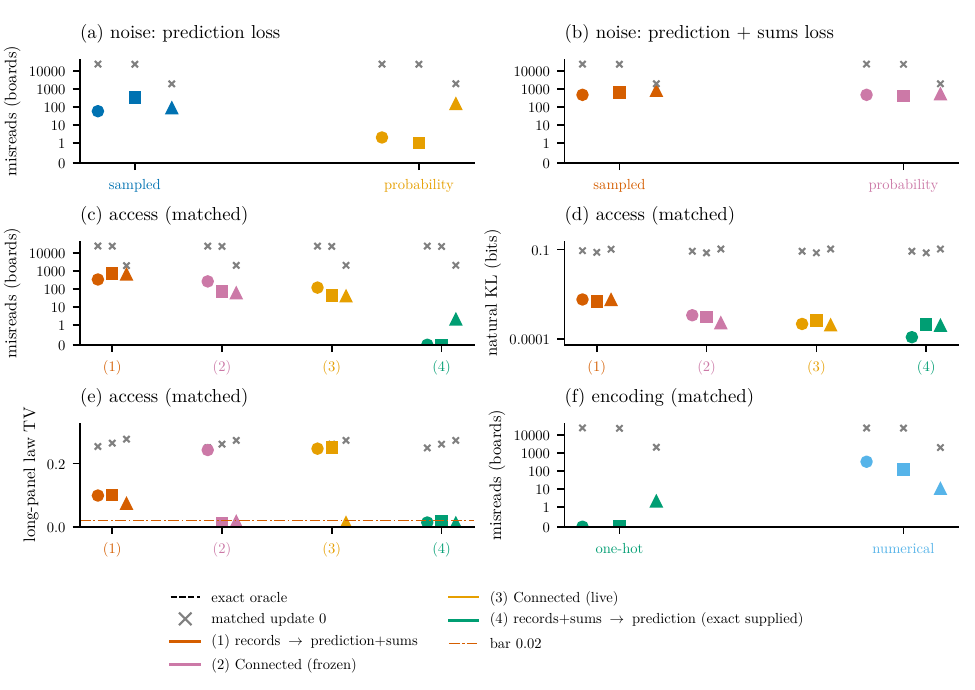}
\caption{\csname captionfig4\endcsname}\label{fig:fig4}
\end{figure}

\textbf{Claim: the noise of sampled targets is one contributing condition.}
In this matched r13 comparison, training on the exact law instead of one
sampled next category removes target noise and changes nothing else. For records $\to$ prediction, census misreads
go from 58 / 328 / 87 with sampled targets to 2 / 1 / 149 with probability
targets: two seeds fall to near zero and the third gets worse
(\cref{fig:fig4}a). For records $\to$ prediction+sums they go from
473 / 648 / 815 to 473 / 400 / 523 (\cref{fig:fig4}b). No run passes under
either target. Removing target noise neither repairs every seed nor produces a pass.

\subsection{Computing or recovering the sums does not guarantee access}\label{sec:res-access}

\textbf{Claim: a network can produce the sums largely correctly, or hold them
in recoverable form, and still not deliver them to the prediction; access is
a separate achievement.}

\emph{Supplied versus learned sums (r12, matched versions).} In r12, the matched
comparison of the declared versions differs in three ways at once: whether
exact sums are supplied, how accurate the sums are, and whether the sums
are supervised. The records+sums $\to$ prediction
networks, which receive exact sums, pass 2 of 3 runs with 0--1 misreads. The
records $\to$ prediction+sums networks, which are trained to output the sums
but do not feed them to the prediction, pass 0 of 3 with 473--815 misreads.
Among boards where their own sums output gets the slot-1 category right,
their prediction still has a categorical misread on 326 / 617 / 770. Their sums output is
correct on 19,269 / 20,516 / 19,544 of the 24,435 completed sum boards, so
these networks compute the sums largely but not exactly. The comparison
supports more reliable prediction with exact supplied sums under these
conditions; it does not isolate access.

\emph{Connection (r13, matched).} The direct test of access is r13's
connection experiment, which connects the learned sums output to the
prediction and changes nothing else (\cref{fig:fig4}c--e). Relative to the
disconnected networks (331 / 725 / 637 misreads), Connected (frozen) has
259 / 69 / 58 misreads, a 1.3--11-fold reduction, and Connected (live) has
117 / 43 / 39, a 2.8--17-fold reduction. Natural KL falls as well, from
0.0018--0.0021 bits to 0.0003--0.0006 bits. Access does not by itself produce reliable prediction: no connected run passes every criterion,
and long-history failures can remain (Connected (live), seed 1: long-panel
maximum 0.251). This supports access as a cause of the misread reduction,
not a selective internal representation of the sums.

\emph{Swap test (r10).} In r10, probes recover all five sums and the swap
test passed calibration,
but on the trained networks the tested alignment was not selective: 188--436
of 1,024 swaps changed another slot's decoded sum. The failed selectivity means only that this test cannot support use under
this alignment.

\subsection{The form of access matters}\label{sec:res-encoding}
\begin{figure}[tb]
\centering
\includegraphics[width=\textwidth]{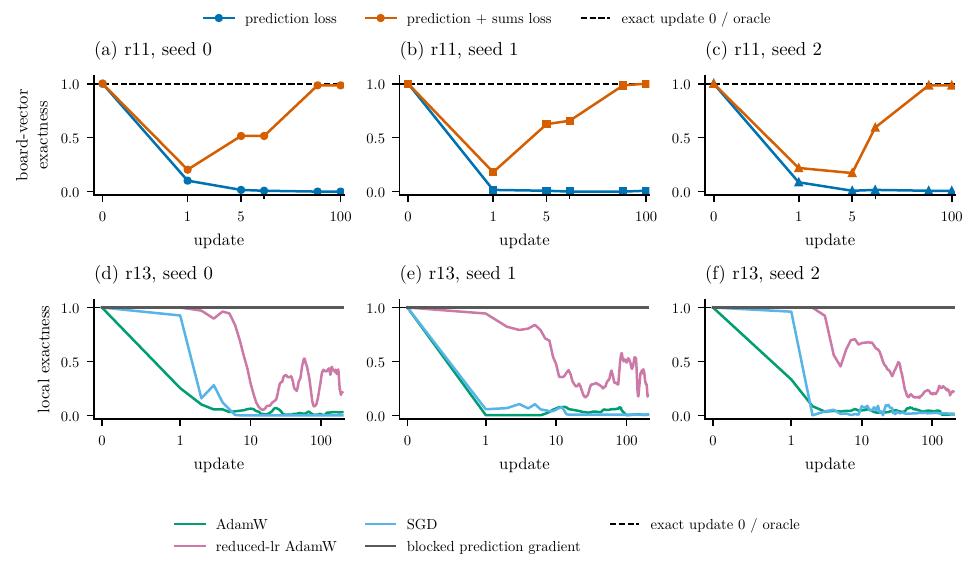}
\caption{\csname captionfig5\endcsname}\label{fig:fig5}
\end{figure}

\textbf{Claim: the same exact information can be easy or hard to use,
depending on how it is encoded.} Given the same exact sums, the one-hot
encoding gives 0 / 0 / 2 misreads and the numerical encoding
$(s/36,\,1-s/36)$ gives 317 / 121 / 10 (\cref{fig:fig4}f). The two
conditions differ only in the encoding. A candidate reason, which these runs
do not test, is that a numerical input requires the prediction to place a
threshold between adjacent values such as 12 and 13 twelfths, whereas a
one-hot input gives each value its own input unit. A supplementary probe analysis of the cutoff values is in
\cref{app:probes}. We do not infer a precision ceiling from these results.

\subsection{Prediction updates erode an exact sums computation}\label{sec:res-erosion}

\textbf{Claim: a computation that is exact can be lost when training
continues on the prediction alone; preservation is a separate achievement.}
In r11, training continued from an exact sums network. With the prediction
loss only, board-vector exactness on a fixed panel of 128 boards falls from
1.0 at update~0 to 13/128, 2/128 and 11/128 at update~1 and remains
severely degraded at later sampled readouts
(\cref{fig:fig5}a--c). Keeping the sums loss largely protects the
computation, though not everywhere: at the endpoint, 24,434 / 24,435 /
24,419 of the 24,435 completed sum boards are correct.

r13 measured the first 200 updates exhaustively on the 1,555 local
histories (\cref{fig:fig5}d--f). Exactness is lost at update~1 under every
tested active optimizer, with very different severity. After one update,
AdamW keeps 393 / 5 / 518 histories exact, AdamW with a tenfold smaller
learning rate keeps 1,554 / 1,468 / 1,553, and SGD keeps 1,440 / 89 / 1,495.
When the prediction gradient to the sums network is blocked and weight decay
continues, all 1,555 histories stay exact through 200 updates in every seed.
Within the range of cumulative movement that all three optimizers covered,
exactness falls for each (\cref{fig:appendix_D_movement}). Erosion is therefore not
specific to Adam. These runs do not separate the direction of the prediction
gradient, the size of the displacement and the margins of the sums answers
as causes.

\subsection{Correct sums do not guarantee reliable prediction}\label{sec:res-reliable}

\textbf{Claim: even with exact sums supplied, reliable prediction is a
further achievement that some seeds miss.} With exact sums supplied as
input, r8's records+sums $\to$ prediction passes 3 of 3 runs, and records+sums $\to$ prediction (sums route only) passes 2 of 3. r12's records+sums $\to$
prediction passes 2 of 3, and r13's one-hot exact condition passes 1 of 3
(\cref{tab:table2}). In the all-slots task (r10), with its own criteria, the
version given exact sums passes 0 of 3. These are historical
comparisons across studies with different training protocols. A pass is also
not exhaustive correctness: the passing r13 seed still misreads 2 boards in
the census. The failed criteria of the supplied-sums networks are specific: r12 seed~2
exceeds the bar on the reset swap (TV 0.0590), r13 seed~0 on the single
\Lcat{} history (law TV 0.02275), and r13 seed~1 on the neutral swap (TV
0.02067). The sums are exact in all of them; candidates include the history update, the probability readout and the
response to interventions. We have not identified their mechanisms.

\section{Discussion}\label{sec:discussion}

\paragraph{What we learned.} Training a network on an upper theory gave it
the lower theory only in a jagged form, and that jagged form is where its
jagged competence came from. The network did learn something of A: the part
of A that B depends on became readable, while the parts B ignores became less
readable. But
what it learned was uneven in the places that matter for B. Category misreads
concentrated near the slot-1 cutoffs, where B's categories change, even
though a probe could read the slot-1 sum on 97--98\% of boards;
when A was trained as an explicit output it was largely but not exactly
right; and part of the prediction's failure sat where this partial A failed.
Seen this way, jagged competence is not noise on top of an otherwise correct
solution. Uneven acquisition, access and preservation of A explain part of
B's jagged competence; some failures remain unexplained. When B instead needed all five sums
(the all-slots task), all five became readable, which suggests that the
jagged form of A follows what B asks of it; that comparison is historical
and crosses tasks.

\paragraph{Knowing an intermediate is not one thing.} Our second lesson is
that ``the network knows A'' hides five separate facts. A can be readable
without established exact computation, computed without reaching the
prediction, delivered in a form that supports less reliable prediction, or
lost under further training, and an exact, delivered A can still leave
failures in B. Matched comparisons isolate the effects of connection,
encoding and prediction gradients: connecting a learned A to B cut misreads
by up to 17-fold, the same exact A gave fewer misreads as one-hot inputs
than as numerical inputs, and a single B update made an exact A inexact unless the
B gradient was blocked. Having the right information is therefore not enough;
how it reaches the decision, in what form, and whether it survives training
all matter. Nor is more supervision of the intermediate automatically
better: networks also trained to output the sums had more misreads (473--815)
than networks trained on the prediction alone (58--328) in r13's
sampled-target condition (\cref{sec:res-noise}).

\paragraph{Jaggedness has locatable sources.} The known layers let us
distinguish several sources of failure: missing access, missing
preservation, a less usable encoding, noisy targets, and history coverage,
assessed in a historical comparison. Some cases remain unexplained: 19 of the 42 diagnosed
r12 histories that exceed the diagnostic bar, and the residual failures with
exact sums. Average error and probe accuracy each looked healthy while these
failures were present, so neither alone establishes complete, reliable
realization of the task. ``Jagged'' names the phenomenon; this account, not the
word, is the explanation.

\paragraph{Relation to other accounts.} Coverage-based accounts explain
jaggedness by gaps in what a system has seen \citep{gans2026}; history
coverage is one such factor here, but the access, preservation and encoding
effects appear in matched comparisons with identical training data, so
coverage alone does not account for them. Other studies report gaps between
what probes read and what outputs do \citep{basu2026,darade2026}; our
connection experiment shows one way such a gap can cause errors and be
reduced.

\paragraph{What we expect to carry over.} We expect the two measurement
cautions to hold wherever the same measurements are used: low average error
does not establish reliability on constructed histories, and recoverability
does not establish predictive use. Whether learning an upper theory yields a
jagged lower theory in larger systems is a test to run, for example by
checking which parts of a known intermediate become readable under
final-output training, whether a learned intermediate output is used by the
main prediction, and whether downstream training preserves it.

\section{Limitations}\label{sec:limitations}

\begin{enumerate}
\item \emph{Scope.} One task family, small recurrent networks (about 40,000
  parameters), three seeds per condition and fixed budgets: we report
  observed counts, not population frequencies, and cannot claim that the
  separations hold for other tasks, architectures or longer training.
\item \emph{Constructed tests.} The law panels and the census locate failures
  but do not estimate how often they occur under the natural distribution.
\item \emph{Confounded comparisons.} Comparisons across studies (r11 versus
  r12; anything involving r10) change more than one factor and are not
  attributed to a single cause; within r12, the version comparison changes
  supplied sums, sums accuracy and sums supervision together and does not
  isolate access.
\item \emph{Internal structure.} Probe declines show reduced measured
  readability, not removal of information, for the tested probe families and
  support; probe boards are disjoint between fitting and evaluation but not
  necessarily unseen during network training, and rare r10 sums (28--36) lie
  outside the registered scope. The connection experiment changes the route,
  not the internal state, and the failed swap selectivity excludes no other
  representation. None of these establishes how the free networks represent
  or use the sums internally.
\item \emph{Encoding.} The encoding experiment compares two encodings and
  does not establish a precision ceiling; the supplementary cutoff
  discrimination is established on that support, not across all cutoff
  boards, and does not diagnose the numerical-encoding failures.
\item \emph{Mechanisms.} Erosion depends on the prediction gradient, but the
  runs do not separate gradient direction, displacement and answer margins,
  so interference is not established; the residual failures of networks
  given exact sums are located but not explained, and a pass is not
  exhaustive correctness.
\end{enumerate}

\section{Conclusion}\label{sec:conclusion}

We asked where jagged competence comes from. In a task built from two known
layers, the answer was that a network trained on the upper layer learns the
lower one only in a jagged form, readable where the upper layer needs it,
with B's category responses failing near the cutoffs, and not reliably
delivered or kept; these separations explain part of its failures, and some
remain unexplained. Knowing an intermediate quantity
turned out to be several separate achievements, each of which can fail on its
own, and average error and probes each hid the failures.

The practical lesson is that a network's competence on a task built in
layers should not be judged by its average error or by what a probe can read,
but by measuring, layer by layer, what it has acquired, delivered and kept.
Whether learning an upper theory yields a jagged lower theory in larger
systems is a test to run.

\clearpage
\makeatletter
\immediate\write\@auxout{\string\newlabel{main-text-end}{{}{\number\numexpr\c@page-1\relax}}}
\makeatother
\section*{A note on the Six Birds programme}
Six Birds' organizing picture motivates constructing upper computations over
lower-level objects \citep[\S1.1]{tsiokos2026foundations}. Whether
final-output training discovers such a construction is this programme's
empirical hypothesis, not a theorem about trained networks. Foundations I
separates stability, novelty and directionality certificates and keeps them
independent. Our five
achievements are different empirical measurements. Read in the theory's
terms, our central finding is that learning a strict extension induced the
lower theory only partially, and these separations help explain part of the
measured jagged competence; this is an empirical observation in one task, not a test of
the theory's formal certificate claims, and it does not establish that the
networks implement those layers internally.

\section*{Competing interests}
The author also developed Six Birds Theory, which motivates the task's
layered design and learning hypothesis; the empirical results stand
independently of the theory and bear on that application without validating
its formal claims. The author declares no other competing interests.

\section*{AI-assistance disclosure}
The author used AI tools in preparing this work and takes full responsibility
for its content.

\section*{Availability of materials}
The source package contains the manuscript, the claims ledger and evidence
manifest, the scripts that rebuild every number, figure and table, and the
released data files with their schema (\cref{app:compute}). The study code is
available at \url{https://github.com/ioannist/six-birds-ml}.

\nocite{*}
\bibliographystyle{plainnat}
\bibliography{references}
\clearpage
\appendix
\counterwithin{figure}{section}
\counterwithin{table}{section}
\label{appendix-start}
\section{Study history}\label{app:history}
\begingroup\fontsize{8}{9.5}\selectfont\setlength{\tabcolsep}{3pt}\emergencystretch=1em\hyphenpenalty=10000\exhyphenpenalty=10000
\begin{longtable}{@{}>{\raggedright\arraybackslash}p{1.967in}>{\raggedright\arraybackslash}p{1.967in}>{\raggedright\arraybackslash}p{1.967in}@{}}
\caption{Analytical appendix A. Exact recorded counts or decisions; no pooled reader accuracies. Complete support, settings, controls and per-seed records are in the numbered supplementary tables and released data.}\label{tab:compact_A}\\
\toprule
\textbf{study} & \textbf{recorded compute} & \textbf{version/\allowbreak{}review records} \\
\midrule\endfirsthead
\toprule
\textbf{study} & \textbf{recorded compute} & \textbf{version/\allowbreak{}review records} \\
\midrule\endhead
\midrule\multicolumn{3}{r}{Continued on next page}\\\endfoot
\bottomrule\endlastfoot
r8 & CPU-era; processor model not recorded in saved per-run summaries & 11 \\
r9 & CPU-era; processor model not recorded in saved per-run summaries & 1 \\
r10 & GPU 3 /\allowbreak{} CUDA fp32 execution recorded in summaries and host verification; GPU model and driver not recorded in the public study artifacts & 20 \\
r11 & GPU 3 /\allowbreak{} CUDA fp32 execution recorded in summaries and host verification; GPU model and driver not recorded in the public study artifacts & 13 \\*
r12 & GPU 3 /\allowbreak{} CUDA fp32 execution recorded in summaries and host verification; GPU model and driver not recorded in the public study artifacts & 14 \\*
r13 & GPU 3 /\allowbreak{} CUDA fp32 execution recorded in summaries and host verification; GPU model and driver not recorded in the public study artifacts & 44 \\
\end{longtable}
\endgroup

The evidence history is retained in Supplement Table S3. Sampling, training and
criteria for r8--r13 are in Supplement Tables S4, S6, S8, S10, S12 and S14;
their complete endpoint matrices are S5, S7, S9, S11, S13 and S15. A corrected
audit does not change the registered training endpoint. Follow-ups and supplementary
panels have their own scope; comparisons between studies are historical.

\clearpage
\section{Analytical study comparison}\label{app:tables}
\begingroup\fontsize{8}{9.5}\selectfont\setlength{\tabcolsep}{3pt}\emergencystretch=1em\hyphenpenalty=10000\exhyphenpenalty=10000
\begin{longtable}{@{}>{\raggedright\arraybackslash}p{1.475in}>{\raggedright\arraybackslash}p{1.475in}>{\raggedright\arraybackslash}p{1.475in}>{\raggedright\arraybackslash}p{1.475in}@{}}
\caption{Analytical appendix B. Exact recorded counts or decisions; no pooled reader accuracies. Complete support, settings, controls and per-seed records are in the numbered supplementary tables and released data.}\label{tab:compact_B}\\
\toprule
\textbf{study} & \textbf{registered passes} & \textbf{original-law runs} & \textbf{law TV bar /\allowbreak{} support} \\
\midrule\endfirsthead
\toprule
\textbf{study} & \textbf{registered passes} & \textbf{original-law runs} & \textbf{law TV bar /\allowbreak{} support} \\
\midrule\endhead
\midrule\multicolumn{4}{r}{Continued on next page}\\\endfoot
\bottomrule\endlastfoot
r8 & 5 & 9 & 0.02 (576 history endpoints) \\
r10 & 0 & 9 & 0.05 (180 law rows) \\
r11 & 1 & 24 & 0.02 (576 history endpoints) \\*
r12 & 2 & 15 & 0.02 (576 history endpoints) \\*
r13 & 1 & 27 & 0.02 (576 history endpoints) \\
\end{longtable}
\endgroup

\begingroup\fontsize{8}{9.5}\selectfont\setlength{\tabcolsep}{3pt}\emergencystretch=1em\hyphenpenalty=10000\exhyphenpenalty=10000
\begin{longtable}{@{}>{\raggedright\arraybackslash}p{2.000in}>{\raggedright\arraybackslash}p{1.700in}>{\raggedright\arraybackslash}p{1.800in}>{\raggedright\arraybackslash}p{0.700in}@{}}
\caption{Access does not guarantee registered reliability. Each vertically ordered triple is seeds 0/\allowbreak{}1/\allowbreak{}2, never a pooled mean. Compare within study; cross-study comparisons are historical. r8 full history maximum, r12/\allowbreak{}r13 long N3--N8 maximum: bar 0.02. r10 is a different task: 180 law rows, original bar 0.05 (equal bar 0.02). Natural panels: 5,000 episodes  x  12 rounds. Complete per-seed failed bars, census support and labels: Supplement Table S2.}\label{tab:table2}\\
\toprule
\textbf{condition} & \textbf{natural KL (bits)} & \textbf{law TV (maximum)} & \textbf{B passes} \\
\midrule\endfirsthead
\toprule
\textbf{condition} & \textbf{natural KL (bits)} & \textbf{law TV (maximum)} & \textbf{B passes} \\
\midrule\endhead
\midrule\multicolumn{4}{r}{Continued on next page}\\\endfoot
\bottomrule\endlastfoot
r8: records $\to$ prediction & \begin{tabular}[t]{@{}l@{}}0.00029829\\7.047e-05\\0.00021024\end{tabular} & \begin{tabular}[t]{@{}l@{}}0.25\\0.2548\\0.25649\end{tabular} & 0/\allowbreak{}3 \\
r8: records+sums $\to$ prediction (sums route only) & \begin{tabular}[t]{@{}l@{}}0.00033865\\6.7874e-05\\6.5306e-05\end{tabular} & \begin{tabular}[t]{@{}l@{}}0.017721\\0.011403\\0.020775\end{tabular} & 2/\allowbreak{}3 \\
r8: records+sums $\to$ prediction & \begin{tabular}[t]{@{}l@{}}0.00025483\\6.8052e-05\\8.0944e-05\end{tabular} & \begin{tabular}[t]{@{}l@{}}0.018665\\0.0051108\\0.0059933\end{tabular} & 3/\allowbreak{}3 \\
r10: records $\to$ prediction & \begin{tabular}[t]{@{}l@{}}0.016034\\0.019138\\0.016491\end{tabular} & \begin{tabular}[t]{@{}l@{}}0.3219\\0.28963\\0.30706\end{tabular} & 0/\allowbreak{}3 \\
r10: records $\to$ prediction+sums & \begin{tabular}[t]{@{}l@{}}0.032325\\0.033959\\0.035928\end{tabular} & \begin{tabular}[t]{@{}l@{}}0.27737\\0.25519\\0.23295\end{tabular} & 0/\allowbreak{}3 \\
r10: records+sums $\to$ prediction (sums route only) & \begin{tabular}[t]{@{}l@{}}0.0070603\\0.0059933\\0.0057117\end{tabular} & \begin{tabular}[t]{@{}l@{}}0.12242\\0.079414\\0.10517\end{tabular} & 0/\allowbreak{}3 \\
r12: records $\to$ prediction+sums & \begin{tabular}[t]{@{}l@{}}0.0015898\\0.0020044\\0.0022981\end{tabular} & \begin{tabular}[t]{@{}l@{}}0.064965\\0.087776\\0.044799\end{tabular} & 0/\allowbreak{}3 \\
r12: records+sums $\to$ prediction & \begin{tabular}[t]{@{}l@{}}0.00011363\\0.00029505\\0.00025691\end{tabular} & \begin{tabular}[t]{@{}l@{}}0.014557\\0.010587\\0.012218\end{tabular} & 2/\allowbreak{}3 \\
r13: records $\to$ prediction+sums & \begin{tabular}[t]{@{}l@{}}0.0021068\\0.0018252\\0.0020459\end{tabular} & \begin{tabular}[t]{@{}l@{}}0.0985\\0.10016\\0.072752\end{tabular} & 0/\allowbreak{}3 \\
r13: Connected (frozen) & \begin{tabular}[t]{@{}l@{}}0.00062178\\0.0005485\\0.00034049\end{tabular} & \begin{tabular}[t]{@{}l@{}}0.24257\\0.011304\\0.015699\end{tabular} & 0/\allowbreak{}3 \\
r13: Connected (live) & \begin{tabular}[t]{@{}l@{}}0.00031731\\0.00041094\\0.00029062\end{tabular} & \begin{tabular}[t]{@{}l@{}}0.24683\\0.25128\\0.011997\end{tabular} & 0/\allowbreak{}3 \\
r13: records+sums $\to$ prediction (exact supplied) & \begin{tabular}[t]{@{}l@{}}0.00011384\\0.00030667\\0.00028061\end{tabular} & \begin{tabular}[t]{@{}l@{}}0.013305\\0.01569\\0.010661\end{tabular} & 1/\allowbreak{}3 \\
r13: records+sums $\to$ prediction (Numerical encoding) & \begin{tabular}[t]{@{}l@{}}0.00083668\\0.00079477\\0.00054318\end{tabular} & \begin{tabular}[t]{@{}l@{}}0.24067\\0.069823\\0.18902\end{tabular} & 0/\allowbreak{}3 \\*
r8 board-only calibration & 0.013514 & 0.1875 & 0/\allowbreak{}1 \\*
exact task oracle & 0 & 0 & ceiling \\
\end{longtable}
\endgroup

Supplement Tables S16--S19 retain all census counts, sampled support, law cases
and r10 exposure. Categorical misreads and continuous TV errors remain separate.
The r10 all-slots task uses 180 law rows with its own original-law 0.05 bar;
the original task uses 576 constructed history endpoints and a 0.02 bar.
Equal-law runs are controls, not evidence for the original-law pathway claim.

\clearpage
\section{Calibration and swap procedures}\label{app:calibration}
\begin{figure}[tb]
\centering
\includegraphics[width=\textwidth]{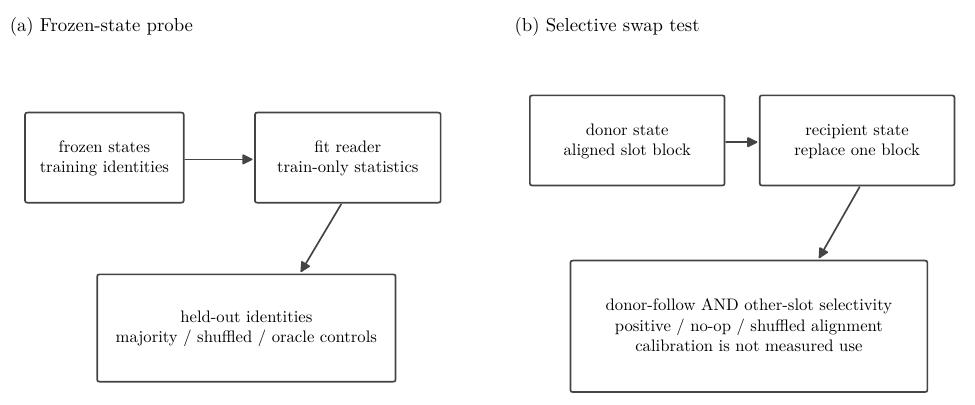}
\caption{\csname captionappendix_C_methods\endcsname}\label{fig:appendix_C_methods}
\end{figure}

\begingroup\fontsize{8}{9.5}\selectfont\setlength{\tabcolsep}{3pt}\emergencystretch=1em\hyphenpenalty=10000\exhyphenpenalty=10000
\begin{longtable}{@{}>{\raggedright\arraybackslash}p{0.843in}>{\raggedright\arraybackslash}p{0.843in}>{\raggedright\arraybackslash}p{0.843in}>{\raggedright\arraybackslash}p{0.843in}>{\raggedright\arraybackslash}p{0.843in}>{\raggedright\arraybackslash}p{0.843in}>{\raggedright\arraybackslash}p{0.843in}@{}}
\caption{Analytical appendix C. Exact recorded counts or decisions; no pooled reader accuracies. Complete support, settings, controls and per-seed records are in the numbered supplementary tables and released data.}\label{tab:compact_C}\\
\toprule
\textbf{study} & \textbf{law} & \textbf{positive /\allowbreak{} null} & \textbf{deciding quantity} & \textbf{pass} & \textbf{support} & \textbf{measured value} \\
\midrule\endfirsthead
\toprule
\textbf{study} & \textbf{law} & \textbf{positive /\allowbreak{} null} & \textbf{deciding quantity} & \textbf{pass} & \textbf{support} & \textbf{measured value} \\
\midrule\endhead
\midrule\multicolumn{7}{r}{Continued on next page}\\\endfoot
\bottomrule\endlastfoot
r8 & original & current\_\allowbreak{}board\_\allowbreak{}null & B deciding criteria & no & 576 & 0.1875 \\
r8 & original & forced\_\allowbreak{}A & B deciding criteria & yes & 576 & 0 \\
r8 & original & forced\_\allowbreak{}raw & B deciding criteria & yes & 576 & 0 \\
r10 & original & exact\_\allowbreak{}oracle & B deciding criteria & yes & 180 law rows & 4.4703e-08 \\
r10 & original & mode\_\allowbreak{}blind & B deciding criteria & no & 180 law rows & 0.48635 \\
r10 & original & untrained & B deciding criteria & no & 180 law rows & 0.42896 \\
r10 & original & slot 1: hard /\allowbreak{} rotated /\allowbreak{} no-op /\allowbreak{} shuffled & donor-follow and selectivity & yes & 1,024 & 1/\allowbreak{}1/\allowbreak{}0/\allowbreak{}0.1123 \\
r10 & original & slot 2: hard /\allowbreak{} rotated /\allowbreak{} no-op /\allowbreak{} shuffled & donor-follow and selectivity & yes & 1,024 & 1/\allowbreak{}1/\allowbreak{}0/\allowbreak{}0.053711 \\
r10 & original & slot 3: hard /\allowbreak{} rotated /\allowbreak{} no-op /\allowbreak{} shuffled & donor-follow and selectivity & yes & 1,024 & 1/\allowbreak{}1/\allowbreak{}0/\allowbreak{}0 \\
r10 & original & slot 4: hard /\allowbreak{} rotated /\allowbreak{} no-op /\allowbreak{} shuffled & donor-follow and selectivity & yes & 1,024 & 1/\allowbreak{}1/\allowbreak{}0/\allowbreak{}0.18164 \\
r10 & original & slot 5: hard /\allowbreak{} rotated /\allowbreak{} no-op /\allowbreak{} shuffled & donor-follow and selectivity & yes & 1,024 & 1/\allowbreak{}1/\allowbreak{}0/\allowbreak{}0.18164 \\
r10 & original & planted contamination & other-slot changes detected & yes & 1,024 & 949/\allowbreak{}950/\allowbreak{}926/\allowbreak{}968/\allowbreak{}959 \\
r11 & original & current\_\allowbreak{}board\_\allowbreak{}null & B deciding criteria & no & 576 & 0.1875 \\
r11 & original & oracle\_\allowbreak{}A & B deciding criteria & yes & 576 & 0 \\
r11 & original & oracle\_\allowbreak{}raw & B deciding criteria & yes & 576 & 0 \\
r12 & original & current\_\allowbreak{}board\_\allowbreak{}null & B deciding criteria & no & 576 & 0.1875 \\
r12 & original & forced\_\allowbreak{}A\_\allowbreak{}oracle & B deciding criteria & yes & 576 & 0 \\
r12 & original & oracle & B deciding criteria & yes & 576 & 0 \\*
r13 & original & board\_\allowbreak{}only & B deciding criteria & no & 576 history endpoints & 0.1875 \\*
r13 & original & oracle & B deciding criteria & yes & 576 history endpoints & 0 \\
\end{longtable}
\endgroup

Deciding-function positive/null comparisons are Supplement Table S20. Table S21
summarizes fixed-pair support, positive validity and other-slot changes; its
named detailed file retains per-slot negative calibration, selectivity and intervals.
Success of an oracle calibration is not evidence of selective use by a trained carrier.
\subsection{Measurement details}\label{app:measurement}

\paragraph{Prediction criteria.} Unseen KL is computed on test positions
whose preceding pair of boards never occurred at a training position; its
support is recorded per run (about 20,000--26,000 positions). Witness
recovery is $1 - \bar{K}/J$, where $\bar{K}$ is the mean KL in bits on the 64
histories made of an anchor and one \Ncat{} board, and $J = 0.061278\ldots$
bits is the Jensen--Shannon divergence between the two laws, the mean KL of a
predictor that ignores the anchor. The state-intervention panels reset and
set the mode, hold it through neutral boards, substitute a board of the same
category and exchange upper states. Under the equal law, witness recovery is
undefined and is replaced by a check that the two witness predictions agree
within 0.02 TV. r10 uses its own conjunction, including bars derived
from its calibrated history-blind predictor and law TV at most 0.05 under
its original law; the complete study-specific criteria for every study,
including r10's covered-value KL, equal-law TV and witness-separation rules,
are listed in Supplement Tables S4, S8, S10, S12 and S14 (r8, r10,
r11, r12, r13; r9 has no pass criteria).

\paragraph{Census.} The two contexts are produced by executing an \Lcat{} and
an \Hcat{} anchor through the network, so a misread can also reflect a
failure to establish the mode. The signature distance is the summed absolute
difference over the two contexts. From r11 on, the census covers all 24,435
boards in each of four fixed renderings and counts are reported per
rendering; r9 used one rendering per board.

\paragraph{Probes.} In the original task the upper state is probed after one
board from a zero state; in r10 it is probed after recurrent histories. The
2,048 probe-training boards and the 512 evaluation boards per slot are
disjoint from each other but not necessarily unseen during network training.
A positive gain is supported when the lower bound of its paired 95\%
interval (1,000 resamples of the same 512 boards) is above zero and, for
linear probes, both fits converged; an interval wholly below zero supports a
decline. Linear probes use balanced class weights, which can push ordinary
accuracy below the majority floor.

\paragraph{Cutoff discrimination (supplementary).} A separate analysis of
the probability-target records $\to$ prediction networks shows that probes
separate the sums 12 from 13 (150 evaluation pairs) and 23 from 24 (13
pairs) in every seed, while those networks still fail registered prediction
tests. This establishes discrimination on that support, not across all
cutoff boards, and does not diagnose the numerical-encoding failures.

\paragraph{Swap test.} For the networks that learn from records, the
alignment is a mean-centred orthogonal map, fitted on training data only,
from the 16-dimensional raw carrier onto 15 coordinates that
track the exact sums (three per slot) plus one residual coordinate; the
patched state is mapped back through its inverse. It is a fixed, fitted
alignment, not a searched subspace. The trained positive (records+sums
$\to$ prediction) is instead patched by replacing the queried slot's hard
sum answer directly. The r10 swap test uses a fixed panel of 1,024 board
pairs per slot and requires at least 20 pairs per slot on which the network
is accurate for both boards. Donor-follow is the fraction of those accurate
pairs for which the patched prediction is closer in TV to the donor's law
than to the base board's law. Selectivity requires that an independent
decoder finds no change in the other four slots' sums. Four checks are
reported separately: whether the trained positive (at the 20,000-update
endpoint) follows the donor, whether the no-op and shuffled-alignment
negatives do not, whether the swap is selective, and whether the evaluated
networks follow the donor.

\clearpage
\section{Learning curves and first-update erosion}\label{app:curves}
\begingroup\fontsize{8}{9.5}\selectfont\setlength{\tabcolsep}{3pt}\emergencystretch=1em\hyphenpenalty=10000\exhyphenpenalty=10000
\begin{longtable}{@{}>{\raggedright\arraybackslash}p{1.967in}>{\raggedright\arraybackslash}p{1.967in}>{\raggedright\arraybackslash}p{1.967in}@{}}
\caption{Analytical appendix D. Exact recorded counts or decisions; no pooled reader accuracies. Complete support, settings, controls and per-seed records are in the numbered supplementary tables and released data.}\label{tab:compact_D}\\
\toprule
\textbf{update-1 condition} & \textbf{correct counts (seeds 0/\allowbreak{}1/\allowbreak{}2)} & \textbf{histories per seed} \\
\midrule\endfirsthead
\toprule
\textbf{update-1 condition} & \textbf{correct counts (seeds 0/\allowbreak{}1/\allowbreak{}2)} & \textbf{histories per seed} \\
\midrule\endhead
\midrule\multicolumn{3}{r}{Continued on next page}\\\endfoot
\bottomrule\endlastfoot
AdamW & 393 /\allowbreak{} 5 /\allowbreak{} 518 & 1,555 \\
reduced-lr AdamW & 1554 /\allowbreak{} 1468 /\allowbreak{} 1553 & 1,555 \\*
SGD & 1440 /\allowbreak{} 89 /\allowbreak{} 1495 & 1,555 \\*
blocked prediction gradient & 1555 /\allowbreak{} 1555 /\allowbreak{} 1555 & 1,555 \\
\end{longtable}
\endgroup

All per-update local counts and movements are in Supplement Table S27.
The supplement also contains every study's learning curves, without checkpoint
selection. The figures below show why low natural KL is not a sufficient reliability
measurement, and the erosion of the sums computation against cumulative
movement.
\begin{figure}[tb]
\centering
\includegraphics[width=\textwidth]{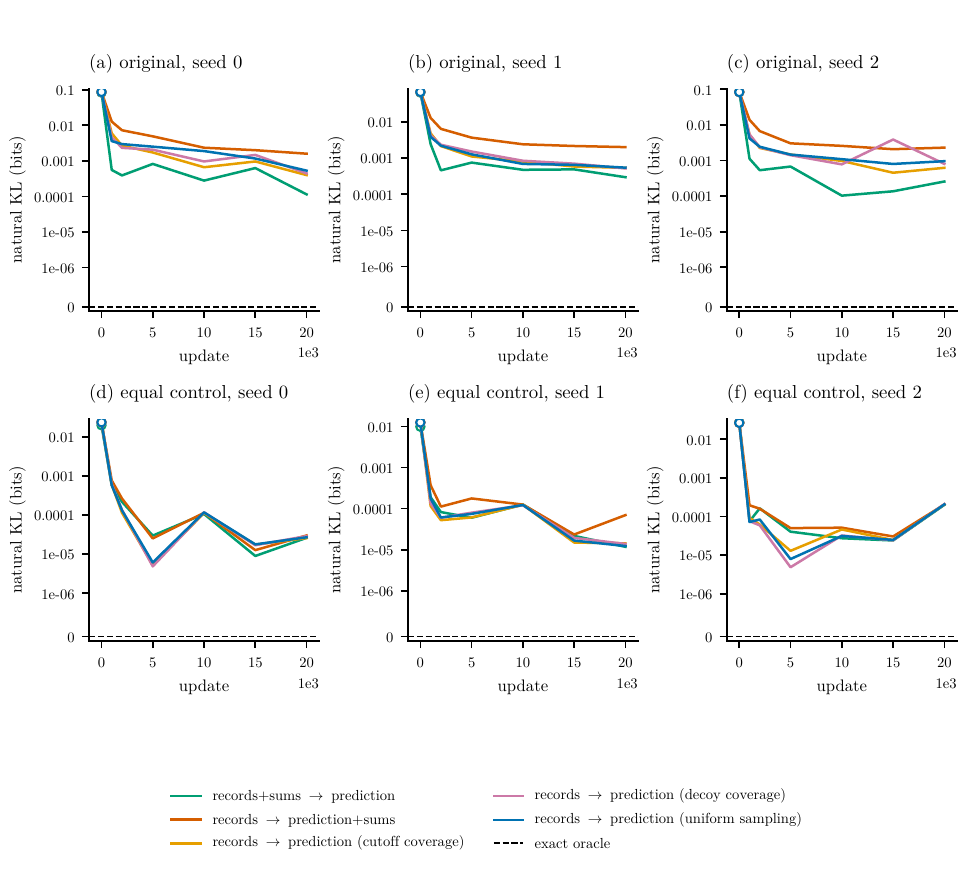}
\caption{\csname captionappendix_D_r12\endcsname}\label{fig:appendix_D_r12}
\end{figure}

\begin{figure}[tb]
\centering
\includegraphics[width=\textwidth]{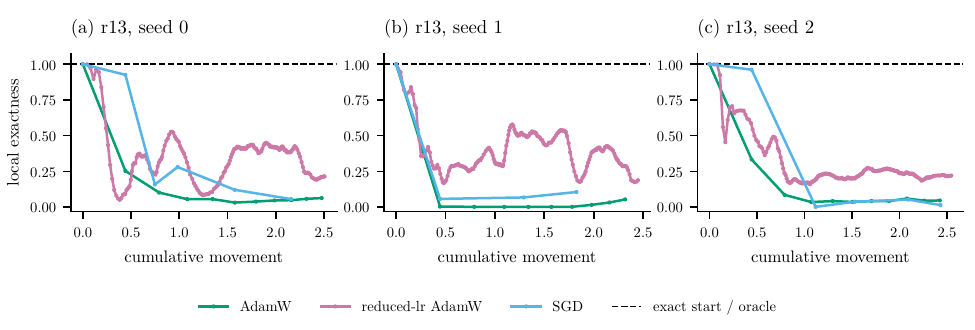}
\caption{\csname captionappendix_D_movement\endcsname}\label{fig:appendix_D_movement}
\end{figure}

\begin{figure}[tb]
\centering
\includegraphics[width=\textwidth]{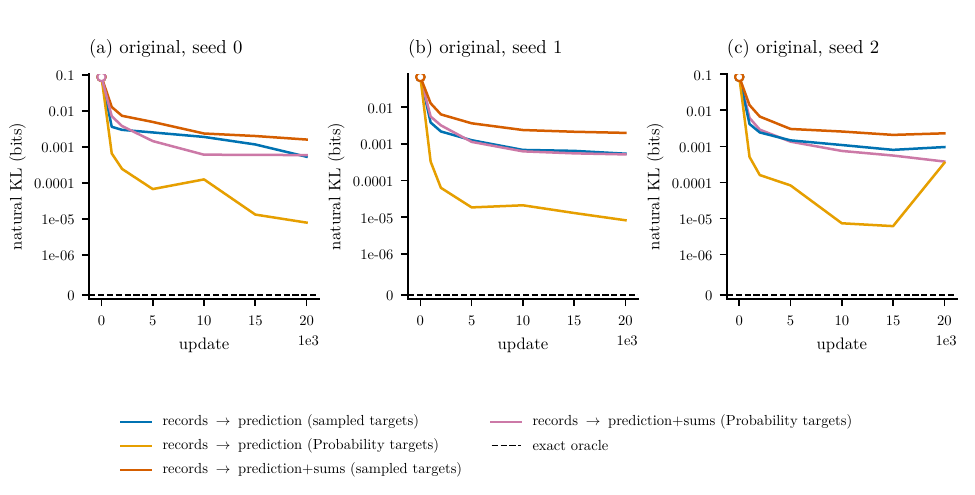}
\caption{\csname captionappendix_D_r13_noise\endcsname}\label{fig:appendix_D_r13_noise}
\end{figure}

\clearpage
\section{Probes and geometry}\label{app:probes}
\begingroup\fontsize{8}{9.5}\selectfont\setlength{\tabcolsep}{3pt}\emergencystretch=1em\hyphenpenalty=10000\exhyphenpenalty=10000
\begin{longtable}{@{}>{\raggedright\arraybackslash}p{0.983in}>{\raggedright\arraybackslash}p{0.983in}>{\raggedright\arraybackslash}p{0.983in}>{\raggedright\arraybackslash}p{0.983in}>{\raggedright\arraybackslash}p{0.983in}>{\raggedright\arraybackslash}p{0.983in}@{}}
\caption{Analytical appendix E. Exact recorded counts or decisions; no pooled reader accuracies. Complete support, settings, controls and per-seed records are in the numbered supplementary tables and released data.}\label{tab:compact_E}\\
\toprule
\textbf{raw linear slot-1 sums reader} & \textbf{seed} & \textbf{correct} & \textbf{boards} & \textbf{gain CI95} & \textbf{converged} \\
\midrule\endfirsthead
\toprule
\textbf{raw linear slot-1 sums reader} & \textbf{seed} & \textbf{correct} & \textbf{boards} & \textbf{gain CI95} & \textbf{converged} \\
\midrule\endhead
\midrule\multicolumn{6}{r}{Continued on next page}\\\endfoot
\bottomrule\endlastfoot
r11 records $\to$ prediction+sums (sums dose 100\%) & 0 & 484 & 512 & 0.5332/\allowbreak{}0.62695 & yes \\
r11 records $\to$ prediction+sums (sums dose 100\%) & 1 & 491 & 512 & 0.60938/\allowbreak{}0.69336 & yes \\
r11 records $\to$ prediction+sums (sums dose 100\%) & 2 & 500 & 512 & 0.68555/\allowbreak{}0.76172 & yes \\
r12 records $\to$ prediction+sums & 0 & 483 & 512 & 0.53315/\allowbreak{}0.62305 & yes \\*
r12 records $\to$ prediction+sums & 1 & 509 & 512 & 0.64648/\allowbreak{}0.72852 & yes \\*
r12 records $\to$ prediction+sums & 2 & 505 & 512 & 0.69722/\allowbreak{}0.77148 & yes \\
\end{longtable}
\endgroup

Complete reader tables, floors, paired intervals and convergence are in Supplement
Tables S22--S24. Original and supplementary geometry support are in S25--S26.
The supplementary geometry figure compares the two recorded conditions of
records $\to$ prediction, sampled and probability targets; its scope is
stated in \cref{app:measurement}.
\begin{figure}[tb]
\centering
\includegraphics[width=\textwidth]{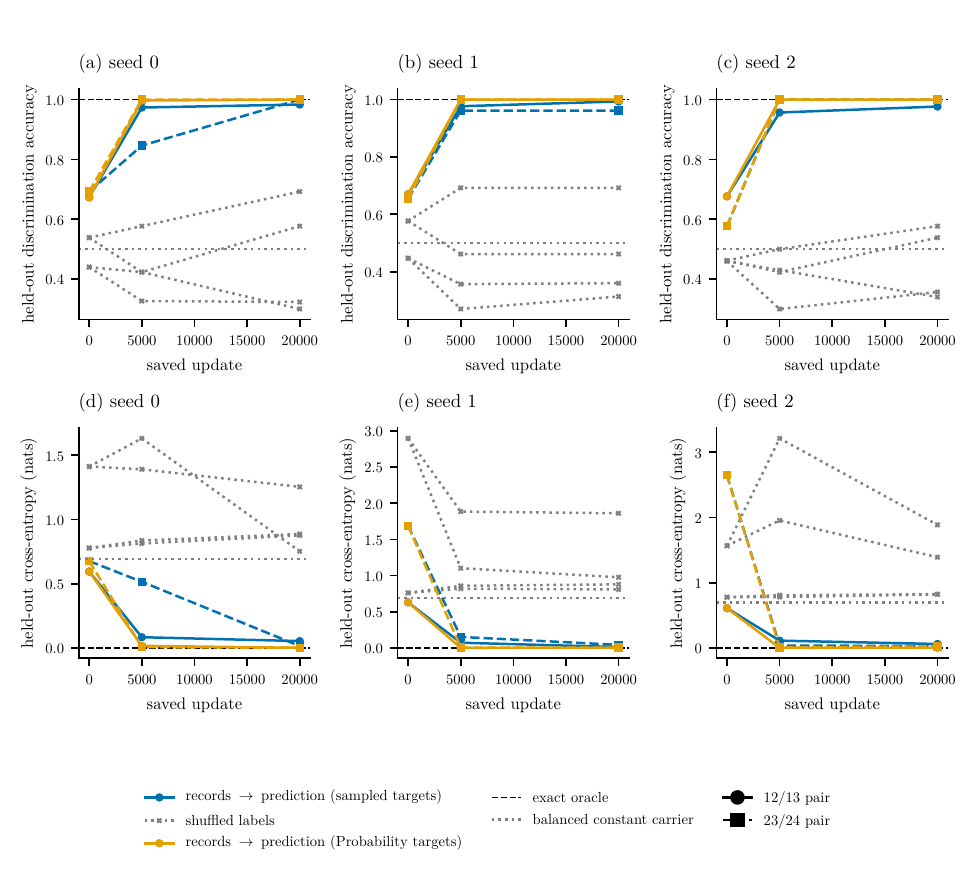}
\caption{\csname captionappendix_E_geometry\endcsname}\label{fig:appendix_E_geometry}
\end{figure}

\clearpage
\section{Compute and reproduction}\label{app:compute}
\begingroup\fontsize{8}{9.5}\selectfont\setlength{\tabcolsep}{3pt}\emergencystretch=1em\hyphenpenalty=10000\exhyphenpenalty=10000
\begin{longtable}{@{}>{\raggedright\arraybackslash}p{1.180in}>{\raggedright\arraybackslash}p{1.180in}>{\raggedright\arraybackslash}p{1.180in}>{\raggedright\arraybackslash}p{1.180in}>{\raggedright\arraybackslash}p{1.180in}@{}}
\caption{Analytical appendix F. Exact recorded counts or decisions; no pooled reader accuracies. Complete support, settings, controls and per-seed records are in the numbered supplementary tables and released data.}\label{tab:compact_F}\\
\toprule
\textbf{study} & \textbf{condition (all seeds)} & \textbf{seconds (0/\allowbreak{}1/\allowbreak{}2)} & \textbf{execution scope} & \textbf{aggregation} \\
\midrule\endfirsthead
\toprule
\textbf{study} & \textbf{condition (all seeds)} & \textbf{seconds (0/\allowbreak{}1/\allowbreak{}2)} & \textbf{execution scope} & \textbf{aggregation} \\
\midrule\endhead
\midrule\multicolumn{5}{r}{Continued on next page}\\\endfoot
\bottomrule\endlastfoot
r13 & AdamW & 21.962/\allowbreak{}19.645/\allowbreak{}20.531 & 200-update erosion; recorded execution & examples, not totals \\
r13 & SGD & 22.219/\allowbreak{}18.44/\allowbreak{}18.835 & 200-update erosion; recorded execution & examples, not totals \\
r13 & blocked prediction gradient & 19.329/\allowbreak{}17.844/\allowbreak{}17.231 & 200-update erosion; recorded execution & examples, not totals \\
r13 & reduced-lr AdamW & 19.241/\allowbreak{}19.387/\allowbreak{}20.737 & 200-update erosion; recorded execution & examples, not totals \\
r8 & records $\to$ prediction & 1855.1/\allowbreak{}1783.3/\allowbreak{}1895.1 & CPU endpoint & examples, not totals \\*
r8 & records+sums $\to$ prediction & 1765.2/\allowbreak{}1810.8/\allowbreak{}1837.4 & CPU endpoint & examples, not totals \\*
r8 & records+sums $\to$ prediction (sums route only) & 1670.6/\allowbreak{}1688.7/\allowbreak{}1682.7 & CPU endpoint & examples, not totals \\
\end{longtable}
\endgroup

Supplement Table S28 summarizes recorded compute quantities; its named detailed
file retains every recorded compute and allocation entry. Unknown
historical execution metadata remain unknown. The public numerical package
contains the manifest, ledger, hash-bound inputs and both TeX entry points.
Rebuild both PDFs from the paper directory with
\texttt{latexmk -pdf main.tex} and, separately,\\
\texttt{latexmk -pdf -jobname=supplement}\\
\texttt{supplement/supplement.tex}.
The package check lists transitive inputs separately for each PDF and refuses
overfull boxes exceeding 10\,pt. Byte-identical asset rebuilds do not imply
byte-identical TeX-engine PDFs, which contain build metadata.
\subsection{Training details}\label{app:training}

\paragraph{Sums interface.} The supplied sums enter the upper state as five
36-way one-hot vectors, projected to 32 dimensions by a bias-free linear map.
The forward pass uses the hard one-hot answer; training gradients pass
through the softmax behind it (a straight-through estimator). The exact sums
network is a learned per-slot update (width 64) that is exact on all 1,555
local histories and is kept frozen except in the erosion continuation.

\paragraph{Episodes.} In r8 and r11, training episodes have up to eight
rounds and omit the third consecutive \Ncat{} board from both input and loss.
From r12 on, each batch holds 128 untruncated nine-round natural episodes and
128 coverage prefixes (an \Lcat{} or \Hcat{} anchor followed by $k$ \Ncat{}
boards, $k=1,\dots,8$, eight prefixes per anchor and length). Coverage
prefixes are scored only at their final position, and the prediction loss is
split equally between natural positions and coverage endpoints.

\paragraph{Rarity and dose.} The rarity conditions of r11 and r12 change only
which board is drawn within a category (25\% enrichment at the cutoff sums or
at matched decoy sums), keeping the category draws and targets paired. The
dose conditions of r11 apply the sums loss on nested 1\%, 10\% or 100\% of
rounds and divide by all active rounds times five slots.

\paragraph{Erosion replays.} The r13 erosion runs last 200 updates and use
r11's truncated episodes, not history coverage. They differ only in how the
sums network is updated: AdamW (learning rate 0.003); AdamW with learning
rate 0.0003; SGD without momentum, whose fixed step size is chosen so that
the norm of its gradient-only displacement on the first shared batch
matches that of AdamW,
with weight decay applied separately (multiplier 0.99997 per update); and a
blocked condition in which the prediction gradient to the sums network is
zero while AdamW weight decay continues. All other parameters use AdamW at
0.003 throughout.

\label{appendix-end}
\end{document}